\documentclass{article}

\usepackage{arxiv}

\usepackage[utf8]{inputenc} 
\usepackage[T1]{fontenc}    
\usepackage{hyperref}       
\usepackage{url}            
\usepackage{booktabs}       
\usepackage{amsfonts}       
\usepackage{nicefrac}       
\usepackage{microtype}      
\usepackage{lipsum}
\usepackage{CJKutf8}
\usepackage{textcomp}
\usepackage{lipsum}

\newcommand\blfootnote[1]{%
  \begingroup
  \renewcommand\thefootnote{}\footnote{#1}%
  \addtocounter{footnote}{-1}%
  \endgroup
}

\title{HG-Caffe: Mobile and Embedded Neural Network GPU (OpenCL) Inference Engine with FP16 Supporting}

\author{
 Zhuoran Ji \\
  The University of Hong Kong, Hong Kong, China \\
  \texttt{jizr@hku.hk}, \texttt{jizhuoran@gsqtec.com} \\
}

\begin{document}

\blfootnote{Documentation of final year project in The University of Hong Kong under the supervision of Prof Wang.}

\maketitle

\begin{abstract}
Breakthroughs in the fields of deep learning and mobile system-on-chips are radically changing the way we use our smartphones. However, deep neural networks inference is still a challenging task for edge AI devices due to the computational overhead on mobile CPUs and a severe drain on the batteries. In this paper, we present a deep neural network inference engine named HG-Caffe, which supports GPUs with half precision. HG-Caffe provides up to $20 \times$ speedup with GPUs compared to the original implementations. In addition to the speedup, the peak memory usage is also reduced to about $80\%$. With HG-Caffe, more innovative and fascinating mobile applications will be turned into reality.
\end{abstract}


\section{Introduction}

In recent years, a great success of deep neural networks has been witnessed, ranging from autopilot cars \cite{bojarski2016end} to wearable heart monitors \cite{shashikumar2017deep} to smart surveillance \cite{liu2016deep}. We attribute the success of deep learning to the advance of algorithms and increased computational power. These deep learning applications are tightly connected to tasks meant for edge AI devices, such as smartphones and embedded devices. Whereas the performance of mobile system-on-chips (SoCs) has advanced by a significant step over the past years, deep neural networks inference is still a challenging task even for the state-of-the-art mobile SoCs \cite{ignatov2018ai}. 

There are many studies designed deep learning frameworks for edge AI platforms. TensorFlow Lite \cite{tang2018intelligent} is a mobile deep learning framework, which supports most common neural network layers. It enables on-devices machine learning inference with low latency and small binary size. Caffe2Go \cite{rodriguez2017intel} is a light-weight and modular mobile deep learning framework, which supports all major generations of hardware and has been deployed on more than 1 billion devices. In addition to these general deep learning frameworks, there are also various dedicate deep neural network inference engines, such as DeepX \cite{lane2016deepx}, Deepmon \cite{huynh2017deepmon}, and Deepsense \cite{yao2017deepsense}.

However, both TensorFlow Lite and Caffe2Go have only CPUs supporting. Executing deep neural network inference on CPUs has a huge computational overhead and will drain out the battery soon. Although few studies offloading the AI tasks to GPUs or DSPs \cite{lane2015can}\cite{latifi2016cnndroid}, as far as we know, no one of them is a general deep neural network inference engine. These programs only support several neural networks layers and hard to extend with new layers.

In this paper, we introduced HG-Caffe\footnote{The Project is avariable on https://github.com/jizhuoran/caffe-android-opencl-fp16.git}, a general deep neural network inference engine, supports all major neural network layers and be optimized for edge AI platforms, such as mobile phones and embedded devices. HG-Caffe supports mobile GPUs, such as Mali and Adreno with both single precision and half precision. As HG-Caffe is a forward only engine, both the peak memory usage and software size are significantly reduced. The main features are summarized as follows:

\begin{enumerate}
    \item GPU supporting with OpenCL
    \item Half precision for all layers
    \item Forward Only Mode
    \item Caffe Weight Files Converter
    \item Third-party library dependencies removed
\end{enumerate}

In the following of this paper, we discuss the meta engineering choices, half precision supporting and report the benchmark results.

\section{Meta Choice}

HG-Caffe is implemented based on BVLC Caffe \cite{jia2014caffe}, with third-party library dependencies removed as much as possible. HG-Caffe has GPU supported with OpenCL, a GPU programming language supported by most mobile GPU architectures. This section discusses these engineering choices in detail.

\subsection{GPU Supporting}

Since AlexNet \cite{krizhevsky2012imagenet} outperformed all the prior competitors significantly in the ILSVRC 2012, GPUs have been widely used in deep neural network training and inference. Most calculations in deep learning are linear algebra, such as matrix multiplication, which involves massive data but simple control logic. For these massive computation tasks, parallel processors, such as GPUs, are more suitable compared to CPUs. The number of ALUs in state-of-the-art mobile GPUs is hundreds of times more than that in mobile CPUs \cite{ignatov2018ai}, which provides ten times throughput. Additionally, lots of state-of-the-art mobile GPUs support native half-precision (FP16) arithmetic \cite{wang2018opencl}. Compare to single precision, FP16 can significantly improve the deep learning throughput because of the reduced arithmetic cost and lowered memory bandwidth pressure \cite{micikevicius2017mixed}.

Furthermore, GPUs are energy efficient processors. Nowadays, TOPs/W becomes essential criteria for the processor used in edge AI platforms, such as mobile phones and embedded devices, which are powered by batteries. Compare to CPUs, GPUs has higher TOPs/W due to the relatively simple control unit and lower frequency. Meanwhile, in the future, the edge AI platforms will use SoC \cite{whitepaper2018}, where the CPU is usually reserved for other usages. For example, on drones, the CPU may need to deal with critical tasks, such as balance maintaining and collision detection, which are time-sensitive, if not real-time required. On the other hand, there may be a deep neural network inference engine used to hand gesture recognition, which needs considerable computations. These less critical takes can take a large proportion of the CPU time if executed on CPUs. Offloading the assistant deep neural networks from CPU to GPU benefits the whole system.

Finally, GPUs are more portable regarding both compilation and performance. Most of the mobile GPUs support Open Computing Language (OpenCL), which is a framework for writing programs that execute across heterogeneous platforms consisting of GPUs. Programs written in OpenCL can be executed on any devices as long as OpenCL standard is supported. Meanwhile, unlike DSP or even dedicated neural network engine, the architectures of GPUs are quite similar among different vendors. The similarity of architectures guarantees the performance of the programs will not differ a lot among various devices.

\subsection{Modified from Caffe}

We implemented our deep neural network inference engine based on BVLC Caffe \cite{jia2014caffe}, a famous deep learning framework with great expression, speed, and modularity. Like other frameworks, Caffe supports most common deep neural network layers. Meanwhile, new layers can be easily integrated into Caffe because of the modularity. 

In addition to the advantages mentioned above, Caffe has better portabilities than other frameworks. AI platforms use various operating systems, ranging from Linux to Android to dedicated operating systems. The deep neural network inference engine needs to be written with a relatively low level and popular language. Caffe is a pure C++ program, which is supported by lots of architecture, even embedded ones.

Furthermore, Caffe is very efficient and lightweight. In the forward phase, Caffe did the least computation among the most famous deep learning frameworks. As Caffe adopts static computation graph, it does not need to track the information necessary for backward. Furthermore, most of the library dependencies required by Cafee ban be turned off or replaced with a small piece of the codes. The less the library dependencies, the easier to deploy the software, especially on dedicated operating systems.

\section{GPU with Half Precision}

The single precision floating point system has been the mainstay for deep neural network training and inference. However, FP32 is more than enough for low learning usage, especially for inference. Meanwhile, most state-of-the-art mobile GPUs support native FP16 arithmetic, such as Mali \cite{koskela2015using} and Adreno \cite{wang2018opencl}, providing potential speedups in mobile platforms. It inspired us to extended official Caffe with fully FP16 supporting. However, FP16 cannot be used for deep neural network inference directly. In this section, we discuss several tricks and engineering choices when implementing HG-Caffe.

\subsection{Batch Norm Layer}
The representable range of FP16 ([$2^{-24}, 65504$]) is quite limited compared to FP32, which makes overflow and underflow occur frequently. According to our experience, overflow and underflow usually occur during accumulation, especially in the batch norm layer and inner product layer. In these layers, a great number of values are accumulated together, which may exceeds the maximum representable values of FP16. Even worse, in the batch norm layer, square operations are involved. For the deep neural networks deal with images, whose value range from $0$ to $255$, square together with the accumulation leads to overflow in most of the cases.

In addition to the accumulation, the scale value multiplied to the vectors can make the whole vector lost to $0$. During calculating the mean of a vector, we need to multiply the accumulated value with the inverse of the length of the vector. However, the inverse of the vector length will exceed the minimum representable values of FP16, which will make the mean lost to $0$.

To solve this problem, we implemented grouped-mean and shifting within the batch norm layer. If we need to calculate the mean of a large number of data, we first partitioned the data into several groups and calculated the mean of each group separately, after which the mean of these parts is computed. Meanwhile, before square operations, the data is shifted by multiplying a scale value, say $\frac{1}{32}$ for data with large values and $32$ for data with small values, to reduce the risk of overflow further. Finally, the data is shifted back at the proper stages.

\subsection{Scalar Value All in Float}

In most deep learning frameworks, the data types for tensors (arrays) can be declared in float and double. However, nowadays, fewer and fewer bits are used for deep learning. There is hardly any deep neural networks can only be trained with double precision. In HG-Caffe, we only support half and float but remove the supporting of double.

Furthermore, different from the tensors, which can be declared in half and float, the scalar values in HG-Caffe are all declared in the float. As half precision arithmetic is supported only by GPUs, half variables need to be converted to float before computed on CPUs. Meanwhile, scalar calculations usually involve lots of logic operations, which should be computed on CPUs. Keeping scalar values in float benefits both the implementation and performance.

However, when these scalar variables are passed to the GPU functions, they need to be converted to half. The conversion is performed in the math functions rather than in the layers' code. Abstracting the conversion from layers ease the workload of implementing new layers. The programmers do not need to write code for both the float and the half. Instead, they only need to pass the tensors in half or float type and scalar values in float type. This engineering choice significantly improves readability and maintenance.




\subsection{Avoiding Conversion and FP16 Weight Files}

The conversion between float and half is costly: each conversion needs four reads and one write. Meanwhile, the conversion function has complex control logic which needs to be executed on CPUs. The overhead of moving computation between GPUs and CPUs are incredibly high, which consists of two memory copies and two conversions. Therefore, we need to reduce the conversion as much as possible.

In HG-Caffe, all layers involve tensors computation support GPU mode with half precision arithmetic. Only the input and output need conversions if they are not in FP16 data type. In addition to the data, HG-Caffe also supports half precision weight files. In the BVLC Caffe implementation, only single precision and double precision weight files are supported. As weight files usually contain billions of numbers, conversion at each load is quite expensive. Loading half precision weight file further reduce the cost of conversion.

However, the mainstay of deep learning field, especially for training, is in FP32. Most of the existing deep neural network models and pre-trained models are stored in FP32. In this project, we also implemented a converter of Caffe models, with which a Caffe model in FP32 can be converted to an identical FP16 one.

\section{Experiment and Result}

We evaluated HG-Caffe on two platforms from different vendors: Kirin with Mali GPUs and Snapdragon with Adreno GPUs. The detailed configurations are shown in table \ref{devices_conf}. Both platforms have 6GB memory, which is shared by GPUs and CPUs.  Noted that Kirin 970 SoC's GPU has 144 ALUs, while Snapdragon 845's GPU has 256 ALUs.


\begin{table}[h!]
\centering
\begin{tabular}{l|l|l|l|l|l}
\hline
Device                      & SoC            & Main Memory & OS          & OpenCL Version & ALUs in GPU \\ \hline
XiaoMi MIX2S                & Snapdragon 845 & 6GB         & Android 8.0 & 2.0            & 256         \\
Hikey 970 Board & Kirin 970      & 6GB         & Android 9.0 & 2.0            & 144         \\ \hline
\end{tabular}
\caption{Platform Configurations\label{devices_conf}}
\end{table}

On each platform, we benchmarked our deep neural network inference engine with various convolutional neural networks, consisting of LeNet on MNIST, AlexNet and ResNet on Cifar-10, and style transfer network with high-resolution images. We compared the execution time and peak memory usage of CPU with single precision, GPU with single precision and GPU with half precision for each neural network. Meanwhile, for each combination, we test different batch size range from 64 to 512 to 1024. The results are shown in table \ref{xiaomi_result} and table \ref{kirin_result}.

\begin{table}[h!]
\begin{tabular}{l|l|l|l|l|l|l|l}
\hline
\multicolumn{2}{c|}{Neural Network Configuration} & \multicolumn{2}{c|}{CPU with FP32} & \multicolumn{2}{c|}{GPU with FP32} & \multicolumn{2}{c}{GPU with FP16} \\ \hline
Name                        & Batch Size          & Time (s)       & Memory (GB)       & Time (s)       & Memory (GB)       & Time (s)        & Memory(GB)       \\ \hline
LeNet                       & 64                  & 0.267          & 0.296             & 0.041          & 0.253             & 0.026           & 0.162            \\
LeNet                       & 512                 & 2.573          & 1.8               & 0.254          & 1.4               & 0.103           & 0.7              \\
LeNet                       & 1024                & 4.292          & 3.5               & 0.713          & 2.7               & 0.33            & 1.4              \\
AlexNet                     & 64                  & 0.878          & 0.0947            & 0.113          & 0.102             & 0.088           & 0.087            \\
AlexNet                     & 512                 & 7.801          & 0.285             & 0.817          & 0.3               & 0.633           & 0.186            \\
AlexNet                     & 1024                  & 14.905         & 0.492             & 1.626          & 0.5               & 1.255           & 0.3              \\
ResNet-20                   & 64                  & 5.812          & 0.075             & 0.449          & 0.084             & 0.362           & 0.075            \\
ResNet-20                   & 512                 & 45.905         & 0.113             & 3.412          & 0.125             & 2.987           & 0.097            \\
ResNet-20                   & 1024                & 120.372        & 0.161             & 6.815          & 0.166             & 6.018           & 0.131            \\
Style Transfer Net          & 1                   & 6.61           & 2.4               & 7.065          & 0.7               & 6.299           & 0.5              \\ \hline
\end{tabular}
\caption{Benchmark Results on XiaoMi MIX2S\label{xiaomi_result}}
\end{table}

\begin{table}[h!]
\centering
\begin{tabular}{l|l|l|l|l|l|l|l}
\hline
\multicolumn{2}{c|}{Neural Network Configuration} & \multicolumn{2}{c|}{CPU with FP32} & \multicolumn{2}{c|}{GPU with FP32} & \multicolumn{2}{c}{GPU with FP16} \\ \hline
Name                        & Batch Size          & Time (s)       & Memory (GB)       & Time (s)       & Memory (GB)       & Time (s)        & Memory(GB)       \\ \hline
LeNet                       & 64                  & 0.377          & 0.306             & 0.466          & 0.09              & 0.311           & 0.097            \\
LeNet                       & 512                 & 2.334          & 1.8               & 2.357          & 0.098             & 1.477           & 0.107            \\
LeNet                       & 1024                & 3.255          & 3.5               & 4.735          & 0.096             & 2.896           & 0.106            \\
AlexNet                     & 64                  & 1.199          & 0.107             & 1.155          & 0.097             & 0.649           & 0.0866           \\
AlexNet                     & 512                 & 6.287          & 0.261             & 14.037         & 0.122             & 4.587           & 0.088            \\
AlexNet                     & 1024                  & 11.261         & 0.505             & 16.008         & 0.115             & 8.99            & 0.092            \\
ResNet-20                   & 64                  & 4.272          & 0.0867            & 3.619          & 0.096             & 2.224           & 0.086            \\
ResNet-20                   & 512                 & 37.004         & 0.127             & 26.771         & 0.097             & 16.724          & 0.086            \\
ResNet-20                   & 1024                & 79.582         & 0.171             & 56.963         & 0.115             & 33.573          & 0.088            \\
Style Transfer Net          & 1                   & 8.282          & 2.4               & /              & /                 & 32.196          & 0.089            \\ \hline
\end{tabular}
\caption{Benchmark Results on HiKey 970 Development Board\label{kirin_result}}
\end{table}

On Snapdragon 845 SoCs, the inference throughput of GPU (FP32) is up to $20 \times$ higher than that of CPU (FP32). Furthermore, the inference throughput of GPU (FP16) can be twice of GPU (FP32). In most of the cases, the speedup of GPU (FP32) is eight times compared to CPU (FP32), and the speedup is about $16 \times$ with the half precision. Meanwhile, the peak memory usage of GPU (FP32) is only $29.1\%$ compared to CPU ones. With the reduction of the bits number, the peak memory usage of GPU (FP16) is only half of GPU (FP32) and $20.8\%$ of the CPU (FP32). Noted that the speedup of GPU is positively proportional to the batch size. As the GPUs are massively parallel processors, large batch size can fully utilize the hardware. In Style Transfer Net with batch size 1, GPU (FP32) is even slower than CPU (FP32). Even with GPU (FP16), the execution time is only a little shorter than CPU (FP32). However, the increase of batch size bridges the gap between GPU (FP16) and GPU (FP32).

On the other hand, on Kirin 970 SoCs, the GPU (FP32) surpasses the CPU (FP32) only when the neural networks are relatively deep, and the batch size is large. Even for GPU (FP16), the execution time is merely comparable with the CPU (FP32). The reason is that, in Kirin 970, the memory space allocated to GPU computing kernel is limited to around 100 MB, which is too small for deep neural networks. The data needs to be swapped between main memory and disk frequently, causing page thrashing and lead to poor performance.

\section{Related Work}

There have been a number of projects of deploying deep learning application to edge AI platforms: the general deep neural network inference engine from industry, such as TensorFlow Lite \cite{tang2018intelligent} and Caffe2Go \cite{rodriguez2017intel}; and research prototypes with novel features, such as Deep Compression \cite{han2015deep}, DeepX \cite{lane2016deepx} and Deepsense \cite{huynh2017deepmon}.

In late 2017, Google released a light-weight machine learning framework named TensorFlow Lite. TensorFlow Lite is optimized for less powerful devices such as mobile phones and embedded devices. There are many applications based on TensorFlow Lite, such as Mobile Object Detection Neural Style Transfer and Image2Text. On the other hand, Caffe2Go was developed by Facebook which supports a wide range of AI use cases. Caffe2Go support all major neural network layers and various types of hardware. It has been the largest deployments of mobile deep learning with more than 1 billion devices. HG-Caffe was modified from the BVLC Caffe but differ from TensorFlow Lite and Caffe2Go in two aspects. First, HG-Caffe supports mobile GPUs with OpenCL, which provides significant speedup and energy-saving. Meanwhile, HG-Caffe supports half precision in GPU mode and further improve the throughput by up to $2 \times$.

On the other hand, researchers have also demonstrated many mobile deep learning frameworks, provides various novel features. Deep Compression \cite{han2015deep} is a series of techniques aiming to reduce the size of deep neural networks. With pruning, trained quantization and Huffman coding, Deep Compression provides up to $49 \times$ memory reduction, which reduces the overhead of deploying deep neural networks to embedded devices. DeepX \cite{lane2016deepx} has a pair of resource control algorithms, which decomposes monolithic deep model networks into several unit-blocks and performs principled resource scaling. DeepX supports CPUs and GPUs heterogeneously. DeepSense \cite{huynh2017deepmon} is a deep learning framework, which is dedicated to time-series tasks. However, most of these programs are not general deep learning frameworks, which may lack the support of common neural network layers. Meanwhile, as these frameworks usually have different types of weight file, they may not be compatible with the pre-trained deep learning models. HG-Caffe instead is a general deep learning framework, which supports all neural network layers of BVLC Caffe. Furthermore, HG-Caffe adopts the same weight file format with BVLC Caffe and the deep learning models trained with BVLC Caffe can be inference with HG-Caffe.

\section{Conclusion and Future Work}
In this paper, we presented the design, implementation, and evaluation of HG-Caffe: a deep neural network inference engine on GPUs with half precision supporting. HG-Caffe is designed with performance and maintainability in mind and can be used as the underlying engine of various deep learning applications. To support half precision, we proposed and implemented several methodologies regarding both arithmetic problems and practices. HG-Caffe provides up to $20 \times$ throughput with GPUs and even higher with half precision on GPUs. Meanwhile, the peak memory usage is reduced to $80\%$ of the official BVLC Caffe. Future work aims to further improve the performance by fine-tuning the parameters and provides more interfaces.

\bibliographystyle{unsrt}
\bibliography{references}

\end{document}